# 딥 러닝 기반 Android 및 Windows 악성 프로그램 탐지 검토

나즈물이슬람, 신석주*
컴퓨터공학과, 조선대학교

# Review of Deep Learning-based Malware Detection for Android and Windows System

Nazmul Islam, Seokjoo Shin*
Dept. of Computer Engineering, Chosun University
nazmul@chosun.kr, *sjshin@chosun.ac.kr (corresponding author)

## Abstract

Differentiating malware is important to determine their behaviors and level of threat; as well as to devise defensive strategy against them. In response, various anti-malware systems have been developed to distinguish between different malwares. However, most of the recent malware families are Artificial Intelligence (AI) enable and can deceive traditional anti-malware systems using different obfuscation techniques. Therefore, only AI-enabled anti-malware system is robust against these techniques and can detect different features in the malware files that aid in malicious activities. In this study we review two AI-enabled techniques for detecting malware in Windows and Android operating system, respectively. Both the techniques achieved perfect accuracy in detecting various malware families.

## 1. Introduction

The rapid development of technology has enabled increased connectivity among people and machines due to the advancements in wireless communication, social networks, and information systems. With the rise in connectivity, Security concerns and privacy protection remain a major challenge, as cyber criminals develop malwares to attack information systems to access, disclose, disrupt, modify, destruct, or misuse the data with a motivation of financial gain, denial of service or reputation damage [1], [2]. Therefore, these malwares need to be detected and identified to develop defensive mechanisms and provide information security (confidentiality, availability, and integrity) to protect the information and systems. Malwares can be classified into different families with respect to their types, behaviors, and privilege level as shown in Fig. 1. of [2].

Traditionally, malware detection was done by a cyber security expert to analyze and filters different classes of malware, which is time consuming and inaccurate. Recently, Machine Learning (ML) or Deep Learning (DL) enabled technologies are used to perpetuate more severe and advanced attacks that can be undetected by traditional techniques. Therefore, AI-enabled malware detection is necessary to tackle the increased number of such attacks. Many advanced malware detection techniques have been developed such as signature-based, anomaly-based and heuristic-based which uses signature database, virtual environment and ML/DL techniques, respectively [3]. In recent years AI based techniques are starting to replace the traditional malware detection and classification technologies[3], [4]. ML/DL based techniques used for malware detection are depicted in Fig. 1. of [3].

Windows system (computers) and android system (phones) are most targeted by malware attacks due to their larger user base and accessibility [3], [4]. Therefore, in this paper we review two proposed AI-enabled malware detection techniques for windows and android, respectively. Section. 2. discusses the system model of the two proposed methods, followed by Section. 3. which concludes the study, and the final section is the references.

## 2. System Model

The authors in [5] proposed AI-based Malware detection technique that uses both ML and DL algorithm to detect malwares in windows. On the other hand, the authors in [6] proposed DL-bases Siamese Neural (SN) network aided classification technique to detect malwares in android. The detailed system design of the two models is further discussed.

### 2.1. AI-based Malware Detection for Windows System

The authors in [5] propose an Intelligent framework combining feature analysis, data balancing and hybrid learning module for robust malware classification model. The whole system is divided into three layers, data processing layer, data training layer and data testing layer as shown in Fig. 1.

The main objectives of data processing layer are data preprocessing and balancing. An autoencoders (AE) convert the data into a three-dimensional latent space distribution and then Synthetic Minority Oversampling Technique-Edited Nearest Neighbor (SMOTEENN) algorithm is used to balance the dataset using both over-sampling and under-sampling techniques. The balanced dataset is then used for model testing and training phase.

In the training layer, Extreme Gradient Boosting (XGBoost) provided the best result for ML algorithm, which utilized ensemble learning architecture for more accurate and reliable classifier models. For the DL algorithm, AE model was used along with various optimization techniques.

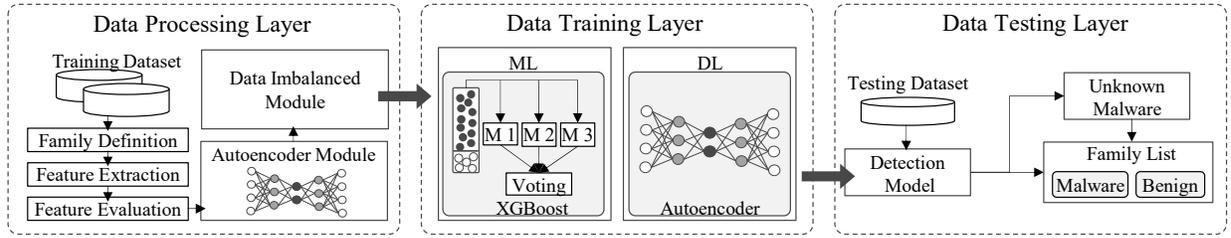

Fig. 1. System architecture of AI-based malware detection in windows system.

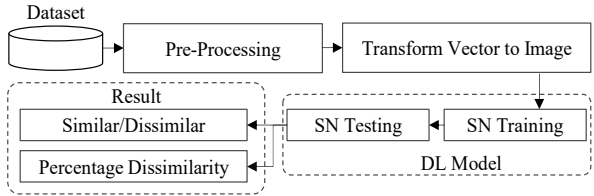

Fig. 2. SN-based malware detection for Android system.

Among them, adaptive moment estimation (Adam) optimization provided the best result. Rectified Linear Unit (ReLU) and Hyperbolic tangent activation function (Tanh) were used in the deep layers, followed by SoftMax in the final layer of the model for non-linearity output. The dataset used in the model was CTU-13 and UNSW-NB15 dataset [7], [8], and among 39 families of malware, 20 were used for training. In the testing layer the model was evaluated on unknown samples for inferencing.

2.2. Malware Detection using SN for Android System.

The authors in [6] used SN Networks to develop DL model for identifying similarities between malicious files in android systems, as shown in Fig. 2. SN consists of two networks which can discriminate between at least two sub-networks and can differentiate between inputs with similar edges or batches. The two networks in SN use the same weights while working in tandem on two different input vectors to compute comparable output vectors. The signature patterns of different malwares in android system can be used to detect and classify them by calculating their similarities using the SN network.

Android programs are compiled into dex (Dalvik Executable) files, then zipped into apk (Android Application Package) which contains code executed by the android runtime. The dex files are divided into three vector regions- Dex File Structure, Index Structural Area, and Data Storage area. The vectors in these three regions are transformed from 8-bit vectors to grayscale images then input in the DL model as images for similarity and dissimilarity score. For the experiment the authors used MALIMG (Malware Image) dataset [9] containing 9,389 grayscale visuals of different dex files originated from 25 different malware families. Performance evaluation of the studies is shown in Table. I.

3. Conclusion

In this study we reviewed two malware detection methods for window and android system. Specifically, we have compared the results of two different malware detection schemes utilizing ML and DL techniques. Both studies suggest that various ML and DL methods can be utilized to achieve exceptional malware detection accuracy, which can otherwise easily outsmart traditional anti-malware systems.

5. Acknowledgement

This research is supported by Basic Science Research Program through the National Research Foundation of Korea (NRF) funded by the Ministry of Education (NRF-2018R1D1A1B07048338).

TABLE I. Comparison between the Deep Learning techniques used for the reviewed methods.

| Platform | DL Models | Advantages | Limitations | AI Performance Analysis (%) | | | | | |
|---|---|---|---|---|---|---|---|---|---|
| | | | | Type | Accuracy | Precision | Recall | F1 | Loss |
| Windows | ML: XGBoost DL: AE | Data balancing and hybrid learning methods | • DL/ML Analysis of only one dataset • No comparison of train and test analysis | ML DL | 99.98 98.88 | 99.94 98.83 | 99.94 98.84 | 99.94 98.94 | |
| Android | SN | Accuracy across different Margin (m) | • No precision, recall and F1 score • No comparison of train and test analysis | DL | 99.67 | - | - | - | 0.5 |